\documentclass{article}
\usepackage{spconf,amsmath,graphicx}
\usepackage{color}
\usepackage{booktabs}
\usepackage{float}
\usepackage{url}
\usepackage{cite}
\usepackage{titlesec}
\usepackage[skip=2pt,belowskip=-12pt]{caption}
\expandafter\def\expandafter\UrlBreaks\expandafter{\UrlBreaks
  \do\a\do\b\do\c\do\d\do\e\do\f\do\g\do\h\do\i\do\j%
  \do\k\do\l\do\m\do\n\do\o\do\p\do\q\do\r\do\s\do\t%
  \do\u\do\v\do\w\do\x\do\y\do\z\do\A\do\B\do\C\do\D%
  \do\E\do\F\do\G\do\H\do\I\do\J\do\K\do\L\do\M\do\N%
  \do\O\do\P\do\Q\do\R\do\S\do\T\do\U\do\V\do\W\do\X%
  \do\Y\do\Z}

\titlespacing*{\section}
{0pt}{1ex}{1ex}

\titlespacing*{\subsection}
{0pt}{1ex}{1ex}

\def\x{{\mathbf x}}
\def\L{{\cal L}}

\title{Mask-RCNN and U-Net Ensembled for Nuclei Segmentation}
\name{Aarno Oskar Vuola \qquad Saad Ullah Akram \qquad Juho Kannala}
\address{Department of Computer Science\\Aalto University}
\begin{document}
%
\maketitle
\begin{abstract}
Nuclei segmentation is both an important and in some ways ideal task for modern computer vision methods, e.g. convolutional neural networks. While recent developments in theory and open-source software have made these tools easier to implement, expert knowledge is still required to choose the right model architecture and training setup. We compare two popular segmentation frameworks, U-Net and Mask-RCNN in the nuclei segmentation task and find that they have different strengths and failures. To get the best of both worlds, we develop an ensemble model to combine their predictions that can outperform both models by a significant margin and should be considered when aiming for best nuclei segmentation performance.
\end{abstract}
\section{Introduction}
\label{sec:intro}

Advances in computer vision algorithms can often be applied to a wide variety of fields, including biomedical imaging. The current widespread use of convolutional neural networks for detection and segmentation tasks has significant applications in the medical field, where tasks often laboriously done by researchers can be replaced by automated systems. Recently, deep convolutional neural networks have seen increased use in biomedical and medical fields in tasks such as organ segmentation from CT scans \cite{nikolov2018deep}. Automatic nuclei instance segmentation from microscopy images is an important task due to the subjectivity of manual segmentations and the increased throughput that data automation enables.  Accurate segmentation requires expert level knowledge and images may contain up to tens of thousands of nuclei that need to be labeled by hand. This seems an ideal application of computer vision, as algorithms can be trained to match experts in accuracy while being able to process thousands of images quickly.

Traditionally, nuclei segmentation has been done with classical computer vision methods such as watershed and active contours. However, neural networks with sufficient amount of training data outperform these systems by a significant margin \cite{Caicedo} and with the increasingly large amount of free and open source software libraries, they have become viable for everyday use in laboratories. For example, a popular open source package CellProfiler now supports adding neural networks to its processing pipeline \cite{mcquin2018cellprofiler}. Object detection and segmentation networks suitable for this task like U-Net \cite{Ronneberger15} and Mask-RCNN \cite{he2017mask} are available as open source libraries, often packaged with pretrained models. Still, tuning these networks to get acceptable results in different domains requires expert knowledge.

While CNNs are an obvious solution to this problem, numerous competing frameworks exist and choosing the best one for different tasks is difficult. This study compares two popular object detection and segmentation frameworks, U-Net and Mask-RCNN, to find where they excel and fail. In addition, an ensemble model combining these two networks' predictions was trained and was found to exceed the performance of both of these models by a significant margin, in some cases more than 5 percent.

Kaggle's 2018 Data Science Bowl \cite{KaggleDSB} presented the nuclei segmentation task in a competition format. The model frameworks used here are inspired by some of the best performing U-Net and Mask-RCNN models from the competition.
\section{Method}
\label{sec:method}

\subsection{U-Net}
\label{sec:unet}

U-Net \cite{Ronneberger15} is a U-shaped convolutional network that uses skip-connections to preserve features at different resolutions. The basic model uses a simple downsampling path, which can be  replaced with a deeper network such as ResNet \cite{He_2016_CVPR}. This allows the model to learn more complex features as the network depth can be greatly increased with residual blocks. Another benefit is that pretrained ResNet networks can be used to initialize the network; for example, pretrained models for ImageNet \cite{imagenet_cvpr09} and COCO \cite{lin2014microsoft} datasets exist. This is especially important when only a relatively small amount of training data is available and the network's learning capacity is increased with a deeper backbone. The backbone selection is largely dependent on the problem complexity and available training data. In this study, the best results were achieved using ResNet101 initialized with weights from a pretrained ImageNet network.

\begin{figure}[t]
\includegraphics[width=\linewidth]{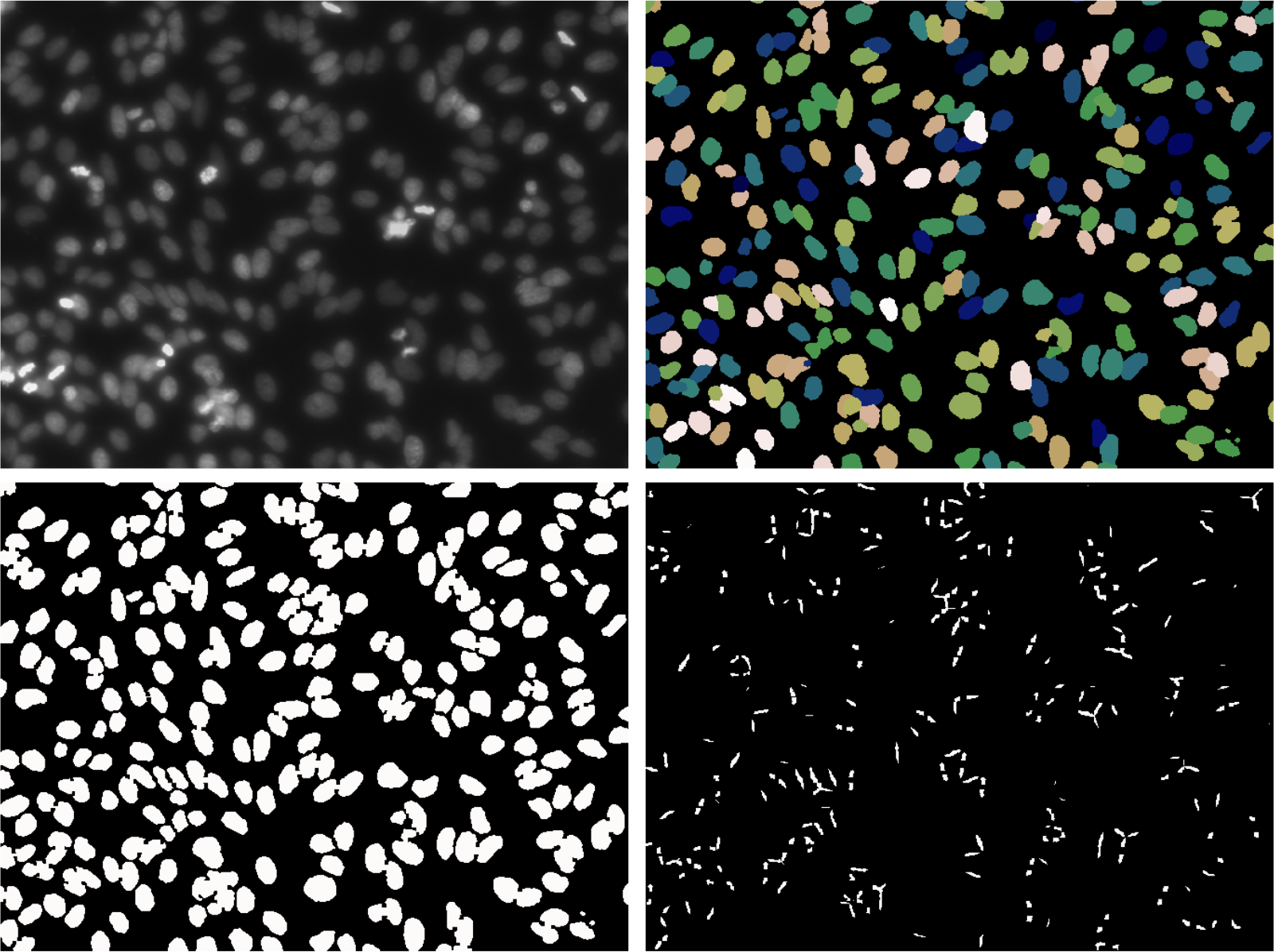}
\caption{Training data and U-Net inputs. Top left: Input image. Top right: Ground truth segmentation mask. Each nucleus is a separate instance. This can be used with Mask-RCNN directly. Lower left: U-Net segmentation mask target. Lower right: U-Net overlapping border target mask.}
\label{fig:unet-targets}
\end{figure}

Instance segmentation is difficult with U-Net, as the output is a binary segmentation mask for the whole input image. Different solutions are available for this problem, such as weighting border pixels heavily in the loss function which was the method used in the original paper \cite{Ronneberger15}. Another approach used by DCAN \cite{chen2017dcan} is to predict the contours of the objects and use post-processing to separate touching instances. Building on this, the method used here is to add an output channel that predicts borders between nearby nuclei, which can later be subtracted from the nuclei segmentation output channel. The targets for the border channel are obtained by dilating the ground truth segmentation and saving the overlapping pixels into a target mask. In addition to helping segmentation, the extra channel should also help the model learn important information about the nucleus shapes. Fig. \ref{fig:unet-targets} visualizes the targets of the U-Net network.

\subsection{Mask R-CNN}
\label{sssec:maskrcnn}

Another popular approach to object segmentation is to use Mask-RCNN \cite{he2017mask} framework. Mask-RCNN is designed to directly address the instance segmentation problem and the effort can then be targeted to tweaking the numerous hyperparameters of the network.

The model predicts bounding boxes for nuclei and then segments the nuclei inside the predicted boxes. While the network is usually able to accurately find bounding boxes for objects, its performance on segmentation seems worse than U-Net's. This is reflected in the results, where Mask-RCNN was found to detect nuclei better but could not segment as accurately.

\subsection{Ensemble model}
\label{sssec:ensemble}
As U-Net and Mask-RCNN frameworks work very differently, an ensemble approach was developed to combine the predictions of both models. A gradient boosting model, similar to one of the top-scoring solutions in the Kaggle competition \cite{Topcoders}, was trained from out of fold predictions of the training data, where structural properties of the prediction mask were used as features and intersection over union (IoU) with ground truth as target. The features used by the ensemble model included structural properties such as eccentricity, perimeter, convex area, solidity etc. calculated with scikit-image's \cite{scikit-image} regionprops function. Adding properties from the input image to the model features was tested but this did not improve the ensemble results. 

At test time, nucleus mask predictions from both U-Net and Mask-RCNN models were used as input to the ensemble model, resulting in IoU with ground truth estimates for both models' predictions. Non-max suppression and thresholding at IoU of 0.3 were then used to construct the final output segmentations. 

The result is that for overlapping masks from U-Net and Mask-RCNN, the mask with highest IoU prediction is used, while for non-overlapping masks they are added to the output segmentation if their predicted IoU is above the threshold. This method was observed to outperform both U-Net and Mask-RCNN, as it was able to combine both models' strengths in different situations. More in depth analysis of performance is done in Sec. \ref{sec:results}.

\section{Experiments}
\label{sec:experiments}

\begin{figure}[t]
\centering
\includegraphics[width=\linewidth]{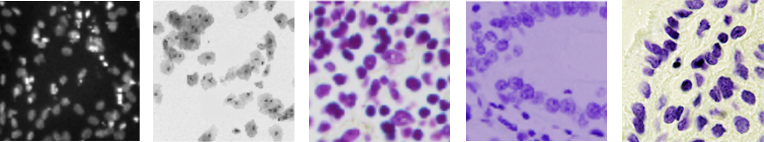}
\caption{Image modalities. From left: Fluorescence, brightfield and three different histology images.}
\label{fig:input_modalities}
\end{figure}

\textbf{Data.}
The goal was to get a well generalized model that could be used with a wide variety of different nucleus images. To this end, training data was gathered from a variety of sources \cite{KaggleDSB,Isbi1,Isbi2,PSB,Murphy,Weebly} that included different types of staining and microscopy techniques. The resulting dataset contained both fluorescence and histology images of a varying quality, to the total of 800 images and masks. Some images were out of focus, had low signal to noise ratio and contained background structures, which all add additional difficulty for prediction. Different image modalities are displayed in Fig. \ref{fig:input_modalities}.

Augmentations were used to increase the training data variance. Augmentations such as shearing, CLAHE, elastic deformations and added noise were found not to work well in this task, leading to the use of only simple augmentations of random flips, rotations, shifts and scaling. Additionally, image contrast transfer \cite{staintools} was used with histology images.
 
\textbf{Training. } 4-fold cross-validation was used to train four models and in the evaluation the results from the four test folds were averaged. Different image classes were balanced in each fold. The models were initialized with ResNet101 backbone, pretrained with ImageNet.

\begin{table}[t]
\footnotesize
\setlength{\tabcolsep}{5.5pt}
\centering
\caption{Performance overview for the models at IoU of 0.7. Over- and undersegmentation marked with oseg and useg.} 
\label{table:results-overall}
\smallskip
\begin{tabular}{lrrrrrr}
\toprule
\textbf{Overall} & mAP & Dice &   Precision &    Recall &  oseg &  useg \\
\midrule
  U-Net & 0.515 & \textbf{0.660} &      0.680 &   0.577 &     52 &    383 \\
  MRCNN & 0.519 & 0.617 &      \textbf{0.812} &   0.596 &     \textbf{14} &    \textbf{164} \\
 Ensemble &  \textbf{0.523} & 0.659 &      0.725 &   \textbf{0.607} &     27 &    328 \\
 \midrule
 \textbf{Fluorescence} & mAP & Dice &   Precision &    Recall &    oseg &  useg \\
 \midrule

 U-Net & 0.564 & \textbf{0.708} &      0.733 &   0.643 &     38 &    301  \\
 MRCNN & 0.569 & 0.684 &      \textbf{0.841} &   0.663 &     \textbf{13} &    \textbf{132} \\
 Ensemble & \textbf{0.570} & 0.703 &      0.767 &   \textbf{0.664} &     23 &    266 \\
 \midrule
 \textbf{Histology} & mAP & Dice &   Precision &    Recall &   oseg &  useg \\
 \midrule
 U-Net & 0.298 & 0.285 &      0.502 &   0.385 &     14 &     82 \\
 MRCNN & 0.300 & 0.236 &      \textbf{0.698} &   0.405 &      \textbf{1} &     \textbf{32} \\
 Ensemble & \textbf{0.316} & \textbf{0.289} &      0.586 &   \textbf{0.442} &      4 &     62 \\
\bottomrule
\end{tabular}
\end{table}

\textbf{Post-processing.} Very small masks (under 10 pixels in area) were removed from predictions and morphological operations were used to fill holes in segmentation masks. To improve U-Net's predictions, a watershed based post-processing method was used to get final segmentation masks for nucleus instances. This was done by restricting the watershed to the areas inside the predicted segmentation mask, using the prediction mask with borders subtracted as markers and finally inputting the mask prediction to the algorithm. Adding the watershed post-processing improved U-Net's detection results by a significant margin, in some cases increasing the mean average precision by more than 0.1. 

As Mask-RCNN segmentation masks may overlap, instances where this happened were resolved by splitting the common region using the distance from each instance.

\textbf{Evaluation.} The metric used in the Kaggle competition \cite{KaggleDSB} was mean average precision (mAP) at different thresholds of the IoU between the ground truth and predicted segmentation.  The thresholds where the precision was calculated were in range [0.5:0.05:0.95], which penalizes errors of only few pixels harshly at the upper end of the range. Here, this same metric is reported along with the object-level Dice coefficient, precision, recall, and number of under- and over-segmentations, i.e. if a model segments a single nuclei into multiple masks or clumps multiple nuclei into one mask.

\section{Results}
\label{sec:results}

\begin{figure}[ht]
\centering
\includegraphics[width=\linewidth]{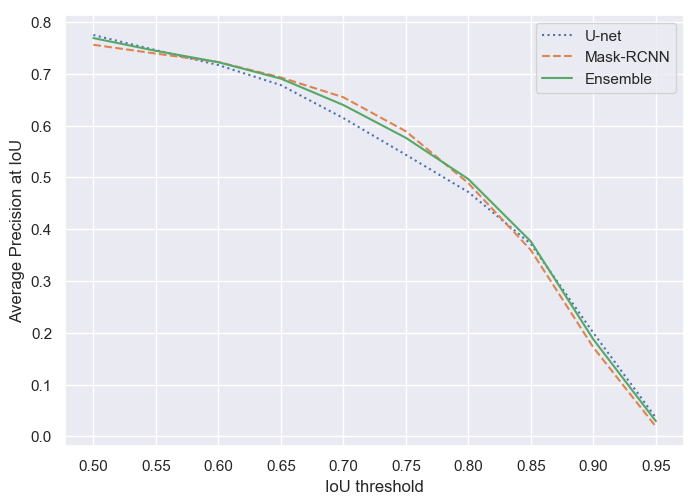}
\caption{AP curve, visualizing the average precision at different IoU thresholds for the models tested.}
\label{fig:ap_curve}
\end{figure}

\begin{figure*}[ht]
\centering
\includegraphics[width=\linewidth]{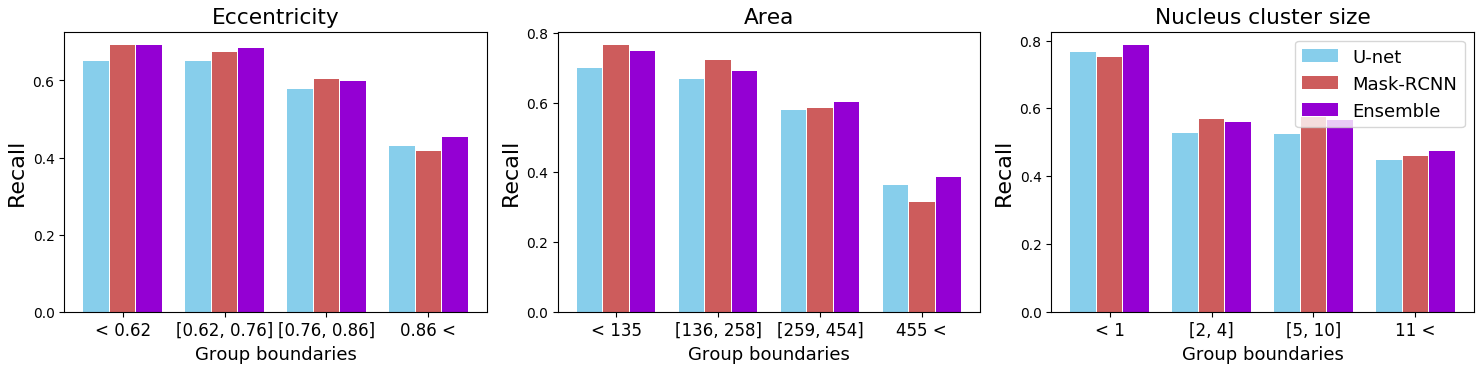}
\caption{Recall at IoU of 0.7 in different scenarios for the different models. Eccentricity, area and cluster size are compared. Each group's boundaries have been set so as to contain the same number of nuclei.}
\label{fig:nuclei-groups}
\end{figure*}

Table \ref{table:results-overall} shows overall performance for the different models. Precision, recall and under- and oversegmentation were calculated at IoU threshold of 0.7 that requires accurate masks while not penalizing errors of only a few pixels. 

\subsection{Overall performance}

Mask-RCNN's and U-Net's differences are highlighted by these results. The mAP of U-Net and Mask-RCNN is quite similar, but the large differences in other metrics reveal their strengths and shortcomings. U-Net's good performance on the Dice score indicates that it is able to create accurate segmentation masks, although with the cost of increased amount of detection errors. Mask-RCNN had worse Dice score, but better recall and precision, indicating that it can detect nuclei more accurately but struggles to predict a good segmentation mask. Another interesting observation is the amount of under- and oversegmentation, where Mask-RCNN had much better performance compared to U-Net. It seems that Mask-RCNN can better detect individual nuclei from a cluster while U-Net has a tendency to clump them into one big nucleus. This is a problem that stems from U-Net's border segmentation output channel, as a very accurate border segmentation with just a few erroneous pixels could result in merged masks. The chosen IoU threshold of 0.7 affects U-Net's recall and precision metric adversely, as its performance was worst in that range.

The ensemble model had the best mAP and recall in all situations. Fig. \ref{fig:ap_curve} plots the average precision at different IoU thresholds, which shows that the ensemble model follows closely the upper bound of both models' results. U-Net had trouble at the mid-range IoU thresholds, which also shows in the slight decrease in the ensemble model's performance. However, Mask-RCNN fared better at the lower and higher IoU thresholds where U-Net produced worse results. The ensemble model did well at picking the best predictions, which shows in the overall good performance across all thresholds.

Both U-Net and Mask-RCNN had worse recall than the ensemble model, but it had slightly worse performance on the precision metric. This is explained by U-Net producing more false positives than Mask-RCNN, some of which leak into the ensemble prediction. This could be prevented by improving the ensemble model's IoU predictions, perhaps with better features. The increase in recall can be attributed to the ensemble model's ability to pick masks from both models' predictions, which leads to increased amount of true positives.

\begin{figure}[!b]
\setlength{\dbltextfloatsep}{0pt}
\includegraphics[width=\linewidth]{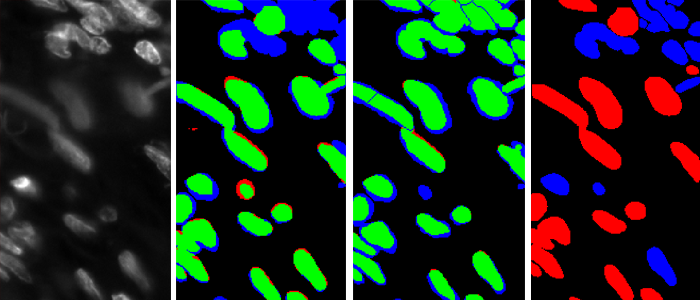}
\caption{Combining the models' predictions on a challenging image. First image is the input to the networks, second and third Mask-RCNN and U-Net predictions respectively. Green pixels overlap with GT, blue pixels are GT with no matching prediction and red pixels a prediction without overlapping GT. Last image shows the ensembled prediction, where U-Net's predictions are marked with blue and Mask-RCNN's with red. mAP increased from 0.28 to 0.32.}
\label{fig:ensemble-nuclei}
\end{figure}

The largest difference between the basic models and the ensemble model is visible when predicting histology images. Both models struggled with this image modality, but again made different errors. This explains why the difference in performance between the basic models and ensemble is large, since good performance can be achieved by picking the best predictions from both models. Fig. \ref{fig:ensemble-nuclei} shows a difficult input image case, where the ensemble model has combined the two models' prediction resulting in increased segmentation accuracy. The result is a combined prediction that is better than either alone.

\subsection{Detailed analysis}

To better understand each of the models' strengths, sensitivity analysis was conducted by evaluating the predictions of nuclei with different structural properties. Fig. \ref{fig:nuclei-groups} visualizes the models' recall in various situations and shows that the ensemble model is at par or better than U-Net and Mask-RCNN in almost every situation. Recall is reported, since a prediction for individual ground truth nucleus can only be classified as a true positive or a false negative depending on if it matches a ground truth nucleus. 

Both models had trouble when nucleus area was large, especially in the group containing all of the largest nuclei. U-Net had better performance here, as Mask-RCNN tended to oversegment the single large nucleus. However, Mask-RCNN made better predictions with small and medium-sized nuclei. When looking at the nucleus size, the ensemble model seems to have the greatest advantage when both U-Net and Mask-RCNN made poor predictions.

The models also had differences when the eccentricity of the nuclei varied. Mask-RCNN had better recall than U-Net, except when the nuclei were very elliptical as Mask-RCNN seems to have trouble creating a good bounding boxes for the nuclei in these situations. Again, the ensemble model had good performance with all eccentricities compared to the basic models.

An important aspect in nuclei segmentation is to look at the segmentation performance when the nuclei are grouped or in close proximity of each other. To find grouped nuclei, the ground truth segmentation mask was dilated and connected components counted. The results show that a lone nucleus is easy to segment, but in the case of multiple clumped nuclei, the network has to learn how to separate the different instances. This is easily seen in Fig. \ref{fig:nuclei-groups}: Prediction quality suffers immediately when the nuclei are clumped together. U-Net fared better when predicting lone nuclei, but Mask-RCNN had slightly better performance on grouped nuclei. U-Net's poor performance on clumped nuclei again stems from the output channel where smallest errors can result in merged masks. The ensemble model had strong performance on all nucleus cluster sizes, indicating that it can accurately pick the best masks from both models even when the nuclei are closely grouped together.

\section{Conclusion}
\label{sec:discuss}
Despite achieving quite similar overall performance on the nuclei segmentation task, the U-Net and Mask-RCNN models made different errors and combining their predictive power using the ensemble model proved to produce better results than either alone. The results indicate that using ensemble model in a nuclei segmentation task improves results, which leads to the question if using it would be beneficial in other instance segmentation tasks as well. Similar instance segmentation tasks in biomedical or medical imaging might see improved performance if an ensemble model would be trained on top of current state of the art solutions. Future work is needed to find if this hypothesis is correct.


\bibliographystyle{IEEEbib}
\bibliography{refs}

\end{document}